%% file: samplepaper.tex
\documentclass[runningheads]{llncs}

\usepackage{graphicx}
\usepackage[colorlinks,urlcolor=blue]{hyperref}

\usepackage[draft]{changes}
\usepackage[noend]{algorithm2e}
\usepackage{multirow}
\usepackage{amsmath}
\usepackage[misc,geometry]{ifsym}

\usepackage{csvsimple}
\usepackage{verbatim}

\usepackage[skip=5pt]{caption}
\usepackage[skip=0pt]{subcaption}
\DeclareMathOperator{\circletool}{circle}

\usepackage{array}

\let\llncssubparagraph\subparagraph
\let\subparagraph\paragraph
\usepackage[compact]{titlesec}
\let\subparagraph\llncssubparagraph

\usepackage{placeins}
\let\Oldsection\section
\renewcommand{\section}{\FloatBarrier\Oldsection}
\let\Oldsubsection\subsection
\renewcommand{\subsection}{\FloatBarrier\Oldsubsection}

\let\Oldsubsubsection\subsubsection
\renewcommand{\subsubsection}{\FloatBarrier\Oldsubsubsection}

\definecolor{greencolor}{rgb}{0,0.5,0}

\newcommand{\maskrcnn}{Mask {R-CNN}\xspace}
\newcommand{\alphapack}{Alpha\xspace}
\newcommand{\betapack}{Beta\xspace}
\newcommand{\gammapack}{Gamma\xspace}
\newcommand{\deltapack}{Delta\xspace}
\newcommand{\epsilonpack}{Epsilon\xspace}
\newcommand{\zetapack}{Zeta\xspace}

\makeatletter
\renewcommand\section{\@startsection{section}{1}{\z@}
                       {-12\p@ \@plus -4\p@ \@minus -4\p@}
                       {8\p@ \@plus 4\p@ \@minus 4\p@}
                       {\normalfont\large\bfseries\boldmath
                        \rightskip=\z@ \@plus 8em\pretolerance=10000 }}

\makeatother

\begin{document}

\title{Learning to solve geometric construction problems from 
images\thanks{This work was partly supported by the European Regional Development Fund under the projects IMPACT and AI\&Reasoning (reg. no. CZ.02.1.01/0.0/0.0/15\_003/0000468 and CZ.02.1.01/0.0/0.0/15\_003/0000466)
and the ERC Consolidator grant \emph{SMART} no.~714034.
}
}

\author{Jaroslav Macke\inst{1,2}\orcidID{0000-0001-6938-4776} \and
Jiri Sedlar\inst{2}\textsuperscript{(\Letter)}\orcidID{0000-0002-4704-3388} \and
Miroslav Olsak\inst{3}\orcidID{0000-0002-9361-1921} \and
Josef Urban\inst{2}\orcidID{0000-0002-1384-1613} \and
Josef Sivic\inst{2}\orcidID{0000-0002-2554-5301}
}

\authorrunning{J. Macke et al.}
\institute{Charles University in Prague, Czech Republic \and
Czech Technical University in Prague, Czech Republic\\
\email{jiri.sedlar@cvut.cz} \and
University of Innsbruck, Austria}
\maketitle
\begin{abstract}
We describe a purely image-based method for finding geometric constructions with a ruler and compass in the Euclidea geometric game. The method is based on adapting the \maskrcnn state-of-the-art visual recognition neural architecture and adding a tree-based search procedure to it. In a supervised setting, the method learns to solve all 68 kinds of geometric construction problems from the first six level packs of Euclidea with an average 92\% accuracy. When evaluated on new kinds of problems, the method can solve 31 of the 68 kinds of Euclidea problems. We believe that this is the first time that purely image-based learning has been trained to solve geometric construction problems of this difficulty.

\keywords{computer vision \and visual recognition \and automatic geometric reasoning \and solving geometric construction problems}
\end{abstract}

\input{section_1_Introduction}
\input{section_2_RelatedWork}
\input{section_3_Euclidea}
\input{section_4_mrcnn}

\input{section_5_unseen_levels}
\input{section_6_experiments}

\input{section_7_conclusion}

\phantomsection
\let\Origclearpage\clearpage
\let\clearpage\relax

\bibliographystyle{unsrt}
\bibliography{bibliography}
\let\clearpage\Origclearpage

\end{document}

%% file: section_1_Introduction.tex
\section{Introduction}
In this work, we aim to create a purely image-based method for solving geometric construction problems with a ruler, compass, and related tools, such as a perpendicular bisector.
Our main objective is to develop suitable machine learning models based on convolutional neural architectures to predict the next steps in the geometric constructions represented as images.

\begin{figure}[tbp]
     \centering
     \begin{subfigure}[b]{0.32\textwidth}
         \centering
         \includegraphics[width=\textwidth]{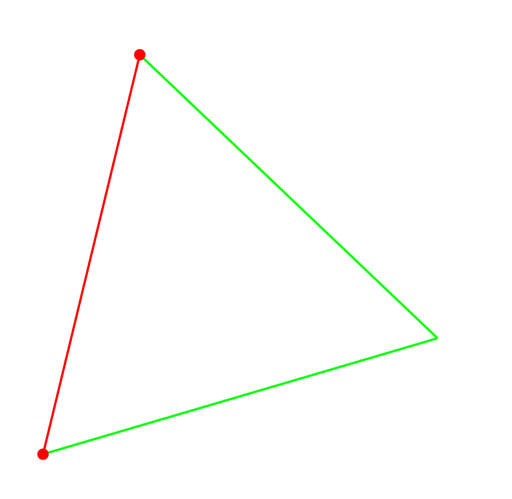}
         \caption{Input}
         \label{fig:euclidea_example_in}
     \end{subfigure}
     \hfill
     \begin{subfigure}[b]{0.32\textwidth}
         \centering
         \includegraphics[width=\textwidth]{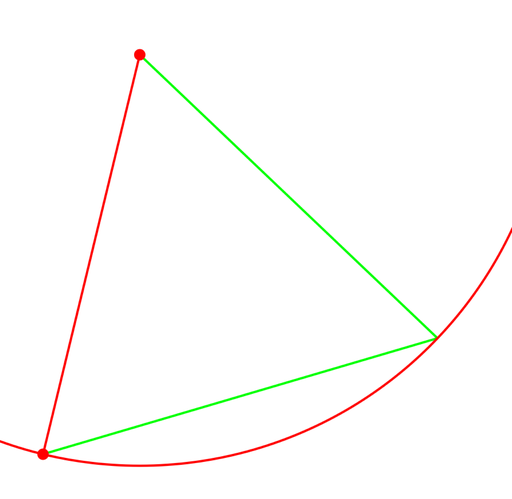}
         \caption{Step 1: Circle tool}
         \label{fig:euclidea_example_s1}
     \end{subfigure}
     \hfill
     \begin{subfigure}[b]{0.32\textwidth}
         \centering
         \includegraphics[width=\textwidth]{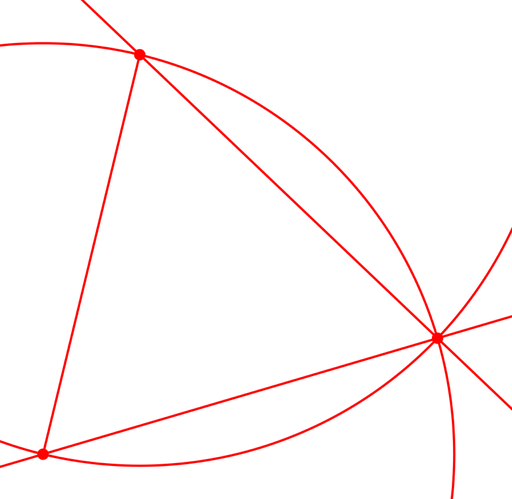}
         \caption{Finished construction}
         \label{fig:euclidea_example_f}
     \end{subfigure}
     \hfill
        \caption{Example solution of Euclidea level \textit{Alpha-05} (construct an equilateral triangle with the given side).
        In all examples, the red channel contains the current state of the construction and the green channel the remaining goal.
        (a) Initial state: one side is given and the goal is to construct the remaining two sides.
        (b) State after the first step: construction of a circle.
        (c) State after the last step: finished construction. 
        }
        \label{fig:euclidea_example}
\end{figure}

In more detail, the input to our neural model is an image of the scene consisting of the parts that are already constructed (red) and the goal parts that remain to be drawn (green).
The output of the neural model is the next step of the construction.
An example of the problem setup is shown in Fig.~\ref{fig:euclidea_example}.

Our first objective is to solve as many geometric construction problems as possible when the method is used in a \emph{supervised setting}. This means solving construction problems that may look very different from images in the training problems, but are solved by the same abstract sequences of construction steps. This setting is still hard because the neural model needs to decide where and how to draw the next construction step in a new image.
Our second objective is to solve problems unseen during the training. This means finding sequences of construction steps that were never seen in the training examples. We evaluate our method in both these settings.

\begin{table}[!htb]
    \centering
    \setlength{\tabcolsep}{6pt}
    \begin{tabular}{l|l}
         Tool & Arguments \\
         \hline
         Point & (coordinates)\\
         Line & (point*, point*)\\
         Circle & (point, point)\\
         Perpendicular Bisector & (point*, point*)\\
         Angle Bisector & (point*, point, point*)\\
         Perpendicular & (line, point)\\
         Parallel & (line, point)\\
         Compass & (point*, point*, point)\\
    \end{tabular}
    \caption{Euclidea tools and their argument types.
    The asterisk denotes interchangeable arguments. For a detailed description of the tools and their arguments see the supplementary material available at the project webpage~\cite{project-page}.}
    \label{tab:tool_overview}
    \vspace*{-2em}
\end{table}

We train and test our models on instances of geometric problems from the Euclidea~\cite{euclidea} game.
Euclidea is an online construction game where each level represents one kind of geometric problem.
Table~\ref{tab:tool_overview} lists the construction tools available in Euclidea. Each level specifies which of the tools can be used.
Euclidea construction problems vary across a wide spectrum of difficulty.
While lower levels are relatively simple or designed specifically to introduce a new tool, more advanced levels quickly grow in difficulty.
These advanced problems are not trivial even for reasonably trained mathematicians, including participants of the International Mathematics Olympiad (IMO).
Our high-level research objective is to explore the question whether computers can learn to solve geometric problems similarly to humans, who may come up with solutions without knowing any algebraic and analytic methods. Solving formally stated IMO problems has already been considered as a grand reasoning challenge\footnote{\smaller \url{https://imo-grand-challenge.github.io/}}.

Solving construction problems from input images poses several challenges. 
First, the same geometric problem can have an infinite amount of different variants with a different scale, rotation or different relative position of individual geometric primitives. The visual solver has to deal with this variability. 
Second, the search space of all possible geometric constructions is very large. For example, a construction with ten steps and (for simplicity) ten different possible construction choices at each step would require searching $10^{10}$ possible constructions. 
To address these challenges we adapt a state-of-the-art convolutional neural network visual recognizer that can deal with the large variability of the visual input and combine it with a tree-search procedure to search the space of possible constructions. We namely build on the \maskrcnn object detector~\cite{maskrcnn} that has demonstrated excellent performance in localizing objects (e.g. cars, pedestrians or chairs) in images and adapt it to predict next steps in geometric constructions, for example, to draw a circle passing through a point in the construction, as shown in Fig.~\ref{fig:euclidea_example_s1}. Despite the success on real images, the off-the-shelf ``Vanilla'' \maskrcnn approach can solve only the very basic level packs of the Euclidea game and adapting \maskrcnn to our task is non-trivial. 
In this work we investigate: (i) how to train the network from synthetically generated data, (ii) how to convert the network outputs into meaningful construction steps,  (iii) how to incorporate the construction history, (iv) how to deal with degenerate constructions and (v) how to incorporate the \maskrcnn outputs in a tree-based search strategy. 

\vspace*{-1mm}
\paragraph{\textbf{Contributions.}}
In summary, the contributions of this work are three-fold.
First, we describe an approach to solving geometric construction problems directly from images by learning from example constructions. This is achieved by adapting a state-of-the-art \maskrcnn visual recognizer and combining it with a tree search procedure to explore the space of construction hypotheses.
Second, we demonstrate that our approach can solve the first 68 levels (which cover all available construction tools) of the geometric construction game Euclidea with 92\% accuracy.
Finally, we show that our approach can also solve new problems, unseen at training. The system as well as our modified Euclidia environment are available online.\footnote{\smaller \url{github.com/mackej/Learning-to-solve-geometric-construction-problems-from-images}, \url{github.com/mirefek/py_euclidea/}} 

The rest of the paper is structured as follows. Section~\ref{sec:related} gives a brief overview of related work.
Section~\ref{sec:euclidea} presents our Euclidea environment.
Section~\ref{mrcnn_section} describes the methods we developed to solve problems in the supervised setting. This includes a description of the neural image recognition methods and their modifications for our tasks.
Section~\ref{sec:unseen_levels} describes our methods
for solving new problems, unseen during the training.
This includes generating sets of proposed steps
and searching the tree of possible constructions. Section~\ref{experiments_section} evaluates the methods on levels seen and unseen during the training.

%% file: section_2_RelatedWork.tex
\section{Related Work}
\label{sec:related}
Visual recognition techniques can be used for interpreting a
geometrical question given by a diagram. Such a geometry solver was proposed in~\cite{allen1,allen2}. The input problem is specified by a diagram and a short text. This input is first decoded into a formal specification describing the input entities and their relations using visual recognition and natural language processing tools.  The formal specification is then passed to an optimizer based on basin hopping.
In contrast, we do not attempt to convert the input problem into a formal specification but instead use a visual recognizer to directly guide the solution steps with only images as input.

The most studied geometry problems are those where the objective is to
find a proof \cite{TGTP}. This contrasts with our work, where
we tackle construction problems. An algebraic approach to a specific
type of construction problem
is used by Argotrics \cite{Argotics}. This is
a Prolog-based method for finding constructions that satisfy
given axiomatically proven propositions.

Automated provers for geometry problems are mostly of two
categories. They are either synthetic \cite{synth1,synth2},
i.e., they mimic the classical
human geometrical reasoning, and prove the problems by applying
predefined sets of rules / axioms (similar triangles, inscribed angle
theorem, etc). The other type of solvers are based on algebraic methods
such as Wu's method \cite{wu} or the Gröbner basis method \cite{grobner}.
These have better performance
than the synthetic ones but do not provide a human readable
solution. There are also methods combining the two approaches. They may for example use some algebra
but keep the computations simple (the Full angle method or
the Area method \cite{AreaBook}). We cannot compare directly with such provers
because of the constructional nature of the problems we study. 
However, our approach is complementary to the above methods and can be used, for example, to suggest possible next construction steps based on the visual configuration of the current scene.

Automated theorem provers (ATPs) such as Otter~\cite{MW97} and Prover9~\cite{McC-Prover9-URL} have been used for solving geometric problems, e.g., in Tarskian geometry~\cite{BeesonW17,DurdevicNJ15,Qua92-Book}. Proof checking in interactive theorem provers (ITPs) such as HOL Light and Coq has been used to verify geometric proofs formally~\cite{BeesonNW19}. Both ATPs and ITPs have been in recent years improved by using machine learning and neural guidance~\cite{JakubuvCOP0U20,GauthierKUKN21}. ATPs and ITPs however assume that a formalization of the problem is available, which typically includes advanced mathematical education and nontrivial cognitive effort. The formal representations are also closer to text, which informs the choice of neural architectures successfully used in ATPs and ITPs (GNNs, TNNs, Transformers).
In contrast, we try to skip the formalization step and learn solving geometric construction problems directly from images.
This also means that our methods could be used on arbitrary informal images, such as human geometry drawings, creating their own internal representations.

%% file: section_3_Euclidea.tex
\section{Our Euclidea geometric construction environment}
\label{sec:euclidea}

Euclidea is an online geometric construction game in 2-dimensional Euclidean space.
The main goal is to find a sequence of construction steps leading from an initial configuration of objects to a given goal configuration.
The construction steps utilize a set of straightedge and compass-based tools (see Table~\ref{tab:tool_overview}).
Every tool takes up to 3 arguments with values specified by the coordinates of clicks on the image of the scene, e.g., $\circletool(A, B)$, where $A$, $B$ are points in the image.
Euclidea is divided into 15 level packs (Alpha, Beta, Gamma, \dots, Omicron) with increasing difficulty; each level pack contains around 10 levels with a similar focus.
In Euclidea, each level has its analytical model, which is projected onto an image and the player does not have access to this model, only to the image.
Each construction is validated with the analytical model to prevent cheating by drawing lines or circles only similar to the desired goal.

In addition, our Euclidea environment can also generate new instances of the levels.
A new instance is generated by randomly choosing initial parameters of the level inside Euclidea.
However, some of these instances can be ``degenerate'', i.e., unsolvable based on the image data.
To prevent such degenerate configurations, we enforce multiple constraints, e.g.,~that different points cannot be too close to each other in the image or that a circle radius cannot be too small.
We use this process of generating new problem instances for collecting examples to train our model.

%% file: section_4_mrcnn.tex
\section{Supervised visual learning approach}
\label{mrcnn_section}
This section describes our method for learning to solve geometric problems.
We build on \maskrcnn~\cite{maskrcnn}, a convolutional neural network for the detection and segmentation of objects in images and videos. Given an image, \maskrcnn outputs bounding boxes, segmentation masks, class labels and confidence scores of objects detected in the input image. 

\maskrcnn is a convolutional neural network architecture composed of two modules. The first module is a region proposal network that proposes candidate regions in the image that may contain the target object (e.g. a ``car"). The second module then, given a proposed candidate region, outputs its class (e.g. ``car", ``pedestrian" or ``background"), bounding box, segmentation mask and confidence score.

\begin{figure}[tbp]
\centering
\includegraphics[width=\textwidth]{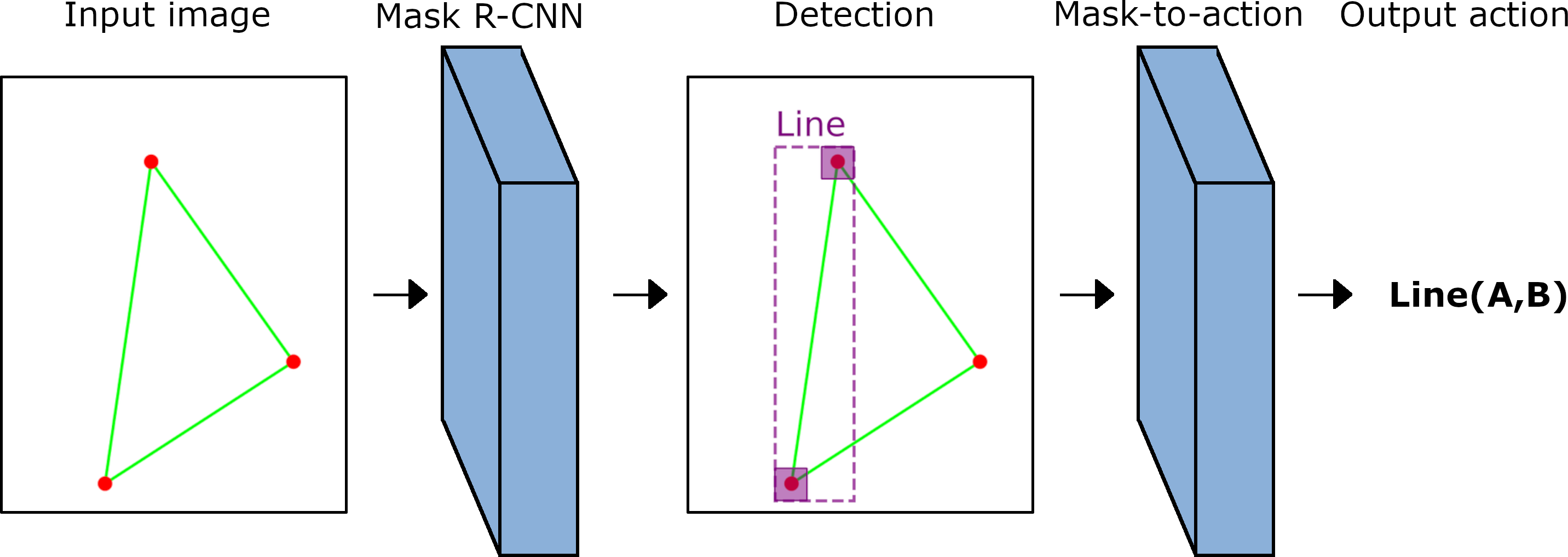}
\caption{Diagram of our approach.
The goal is to construct a triangle given three points.
The input is an RGB image with the current state of the construction in the red channel (the three points) and the goal is given in the green channel (the three sides of the triangle).
The \maskrcnn predicts a line between two of the three points.
The dashed rectangle denotes the bounding box of the line and the small square magenta masks denote the two points on the line.
This \maskrcnn line detection is then converted into a Euclidea tool action, $Line(A, B)$, represented by the tool name and its arguments (the locations of the two points).
The process that converts the \maskrcnn output masks into actions in the Euclidea environment is described in Section~\ref{sec:mask_to_action}.}
\label{our_approach_schema}
\vspace*{-1.5em}
\end{figure}

We adapt the \maskrcnn model for the task of solving geometric construction problems.   
Fig.~\ref{our_approach_schema} shows the diagram of our approach.
The main idea is to train \maskrcnn to predict the tool that should be used at a given step, including its arguments.
For example, as shown in Fig.~\ref{our_approach_schema}, the input is the image depicting the current state of the construction in the red channel of the image (three points in red) and the goal in the green channel (the three sides of the triangle). \maskrcnn predicts here to execute the Line tool. The predicted bounding box of the line is shown in magenta. 
For this purpose, \maskrcnn network has to recognize the two
points in the input image and predict their location, represented by the rectangular masks.
The output masks are then transformed to coordinates of the two points that need to be ``clicked" to execute the Line tool in the Euclidea environment.
 
To train \maskrcnn to solve geometric construction problems, we have to create training data that represent applications of the Euclidea tools and adjust the network outputs to work with our Euclidea environment.
To generate training data for a given Euclidea level, we follow a predefined construction of the level and transform it to match the specific generated level instances (see Section~\ref{sec:euclidea}). Each use of a Euclidea tool corresponds to one example in the training data.
We call each application of a tool in our environment an \emph{action}, represented by the tool name and the corresponding click coordinates.
For example, the Line tool needs two action clicks, representing two points on the constructed line.

The following sections show the generation of training data for \maskrcnn (Section~\ref{sec:action_to_mask}), describe how we derive Euclidea actions from the detected masks at test time (Section~\ref{sec:mask_to_action}), present our algorithm for solving construction problems (Section~\ref{sec:solving_construction}), and introduce additional components that improve the performance of our method (Section~\ref{mrcnn_components}).

\subsection{Action to Mask: generating training data}
\label{sec:action_to_mask}
Here we explain how we generate the training data for training \maskrcnn to predict the next construction step. In contrast to detecting objects in images where object detections typically do not have any pre-defined ordering, some of the geometric tools have non-interchangeable arguments and we will have to modify the output of \maskrcnn to handle such tools.

We represent the \maskrcnn input as a 256$\times$256 RGB image of the scene with the current state in the red channel and the remaining goal in the green channel; the blue channel contains zeros.
Note that for visualization purposes, we render the black background as white.
Each training example consists of an input image capturing the current state of the construction together with the action specifying the application of a particular tool.
The action is specified by a mask and a class, where the class identifies the tool (or its arguments, see below) and the mask encodes the location of each point click needed for application of the tool in the image, represented as a small square around each click location.
The Perpendicular tool and Parallel tool
have a line as their argument (see Table~\ref{tab:tool_overview}), passed as a mask of either the whole line or one click on the line.

\begin{figure}[ht]
     \centering
     \begin{subfigure}[b]{0.32\textwidth}
         \centering
         \includegraphics[width=\textwidth]{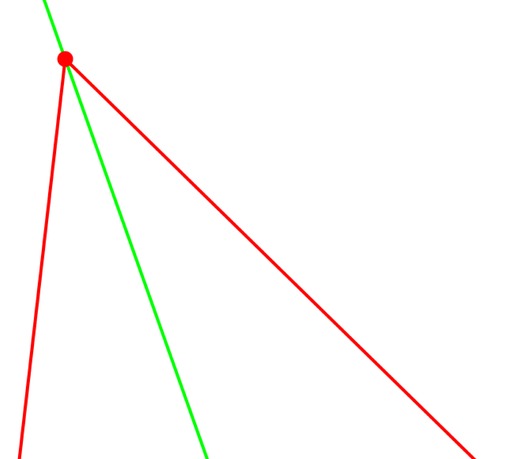}
         \caption{Input}
         \label{fig:training_data_primary_02_02_input}
     \end{subfigure}
     \hfill
     \begin{subfigure}[b]{0.32\textwidth}
         \centering
         \includegraphics[width=\textwidth]{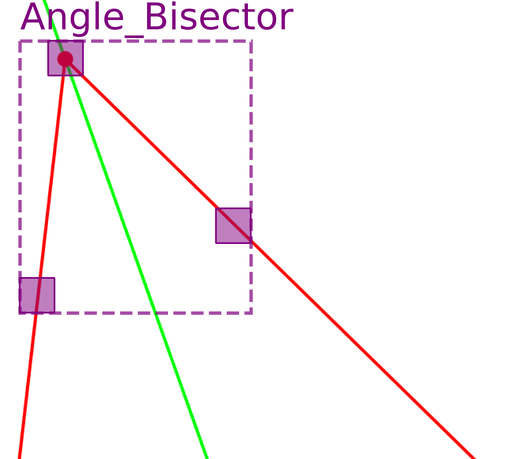}
         \caption{Primary detection}
         \label{fig:training_data_primary_02_02_primary}
     \end{subfigure}
     \hfill
     \begin{subfigure}[b]{0.32\textwidth}
         \centering
         \includegraphics[width=\textwidth]{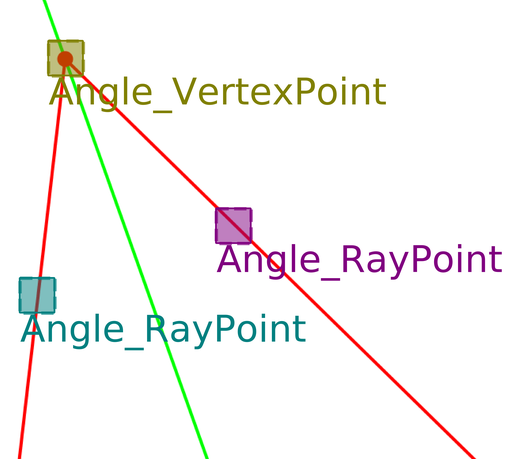}
         \caption{Secondary detection}
         \label{fig:training_data_primary_02_02_secondary}
     \end{subfigure}
        \caption{An example from training data for Euclidea level \textit{Beta-02} (construct the line that bisects the given angle).
        The current state is in red and the remaining goal in green.
        (a) Input for the \maskrcnn model.
        (b) Primary detection of the \maskrcnn model identifying the tool type: Angle Bisector tool (purple).
        (c) Three secondary detections identifying the arguments of the tool: one angle vertex point (yellow) and two angle ray points (purple and turquoise).}
        \label{fig:training_data_primary_02_02}
        \vspace*{-1.5em}
\end{figure}

\paragraph{\textbf{Primary and secondary detections.}}
\label{position_dependent_pars}
Encoding clicks as in the previous paragraph is not sufficient for tools with non-interchangeable parameters.
For example, the Circle tool has two non-interchangeable parameters: 1) the center and 2) a point on the circle, defining the radius.
To distinguish such points, we add a secondary output of the \maskrcnn model.
For example, for the Circle tool, we detect not just the circle itself but also the circle center and the radius point.
We denote the detection of the tool (Line tool, Circle tool, ...) as the \emph{primary detection}, and the detection of its parameters as the \emph{secondary detections}.
Fig.~\ref{fig:training_data_primary_02_02} shows an example of primary and secondary detections for the Angle Bisector tool, including the corresponding classes: Angle\_Bisector (primary detection), Angle\_VertexPoint, and Angle\_RayPoint (secondary detections).
\subsection{Mask to Action: converting output masks to Euclidea actions}
\label{sec:mask_to_action}
To solve Euclidea levels, we have to transform the \maskrcnn output to fit the input of the Euclidea environment.
We refer to this step as ``mask-to-action" as it converts the output of \maskrcnn, which is in the form of image masks specifying the primary and secondary detections (see Section~\ref{sec:action_to_mask}), into tool actions in the Euclidea environment.
The mask-to-action conversion consists of two stages. The first stage obtains locations of individual ``point clicks" from the primary detections for the predicted tool and the second stage determines the order of parameters using the secondary detections.

\paragraph{\textbf{First stage:}} To localize individual points, we use the heat map produced by the final \maskrcnn layer.
The heat map assigns each pixel the probability of being a part of the mask and can be transformed into a binary mask by thresholding.
Instead, we use the heat map directly to localize the detected points more accurately.
We select points with the highest probability in the masks using
a greedy non-maximum-suppression method \cite{non_max_sup}.

\paragraph{\textbf{Second stage:}} We will explain this stage on the example of the Angle Bisector tool (see Fig.~\ref{fig:training_data_primary_02_02}).
A detection of this tool has 4 detection outputs from \maskrcnn, namely, 1 primary and 3 secondary detections.
The primary detection corresponds to the whole tool and the secondary detections to the individual points, i.e., one angle vertex point and two angle ray points (see Fig.~\ref{fig:training_data_primary_02_02}).
To use the Angle Bisector tool, we have to determine the correspondence between the primary and secondary detections.
We obtain 3 point coordinates from the primary detection in the first stage, as described above.
We can also get 3 points from the 3 secondary detections, one point per detection.
Each point in the primary detection should correspond to one point in the secondary detection, but these points may not exactly overlap.
The point correspondence is determined by finding a matching between the primary and secondary points that minimizes the sum of distances between the primary and secondary points such that each point is used exactly once.

\subsection{Solving construction problems by sequences of actions}
\label{sec:solving_construction}
Next, we can create an agent capable of solving Euclidea construction problems.
In the previous section, we have described how to get a Euclidea action from \maskrcnn outputs.
However, \maskrcnn can predict multiple candidate detections (that correspond to different actions) for one input image.
\maskrcnn returns for each detection also its score, representing the confidence of the prediction.
To select the next action from the set of candidate actions (derived from \maskrcnn detections) in each step,
the agent follows Algorithm~\ref{alg:top_score_inference}, which chooses the action with the highest confidence score at each state.
\vspace{-6mm}
\begin{algorithm}[!htb]
\SetAlgoLined
\SetAlFnt{\small\sf}
 \KwResult{Test level inference: True if level completed, False otherwise.}
 Initialize a level\;
 \While{level not complete}{
  $s \gets$ current state of the level\;
  $p \gets model.predict(s)$\;
  \If{predictions $p$ are empty}{
   \Return False\;
   }
  $a \gets$ action from $p$ with highest score\;
  execute $a$
 }
 \Return True\;
\vspace{0.5mm}
\caption{\small{Solving construction problems by choosing the action with the highest score.}}
\label{alg:top_score_inference}
\end{algorithm}
\vspace{-8mm}
\subsection{Additional components of the approach}
\label{mrcnn_components}
Here we introduce several additional extensions to the approach described above and later demonstrate their importance in Section \ref{experiments_section}.
\paragraph{\textbf{Automatic point detection.}}
Our Euclidea environment requires that each point important for the solution is identified using the Point tool.
For example, when we have to find the third vertex of the triangle in Fig.~\ref{fig:euclidea_example}, we have to use the Point tool to localize the intersections of the circles.
The Automatic point detection modification automatically adds points to the intersections of objects.
\paragraph{\textbf{History channel.}}
To better recognize which construction steps have already been done and which still need to be constructed, we add a third, history channel (blue) to the input of~\maskrcnn, containing the construction state from the previous step.
\paragraph{\textbf{4+ Stage training.}}
\maskrcnn is typically trained in 2 stages: first, only the head layers are trained, followed by training of the whole network, including the 5-block convolutional backbone.
The 4+ Stage training modification splits the training into 3 stages: first, the head layers are trained, then also the fourth and fifth backbone blocks, and finally, the whole network.
\paragraph{\textbf{Intersection degeneration rules.}}
To decide whether a generated level can be solved using only the image information, we apply the following rules to identify degenerate configurations: a) the radius of a circle cannot be too small, b) the distance between points, lines, or their combinations cannot be too small.
In this modification, we add a third rule: c) any intersection of geometric primitives cannot be too close to points that are necessary for the construction.
This prevents possible alternative solutions from being too close to each other and the auxiliary intersections created during the construction from being too close to points from the initial state and the goal.
\paragraph{\textbf{On-the-fly data generation.}}
Generating training data on-the-fly allows us to (potentially infinitely) expand the training set and thus train better models.

%% file: section_5_unseen_levels.tex
\section{Solving unseen geometric problems via hypothesis tree search}
\label{sec:unseen_levels}
In the previous section, we have shown how to train a visual recognition model to predict the next step of a given construction from a large number of examples of the same construction with different geometric configurations of the primitives. In this section, we investigate how to solve new problems, which we have not seen at training time.
This is achieved by (i) using models trained for different construction problems (see Section~\ref{mrcnn_section}) to generate {\em a set of hypotheses} for each construction step of the new problem and then (ii) searching the tree of possible constructions. These two parts are described next. 

\subsection{Generating action hypotheses}
\begin{figure}[!htb]
     \centering
     \begin{subfigure}[t]{0.32\textwidth}
         \centering
         \includegraphics[width=\textwidth]{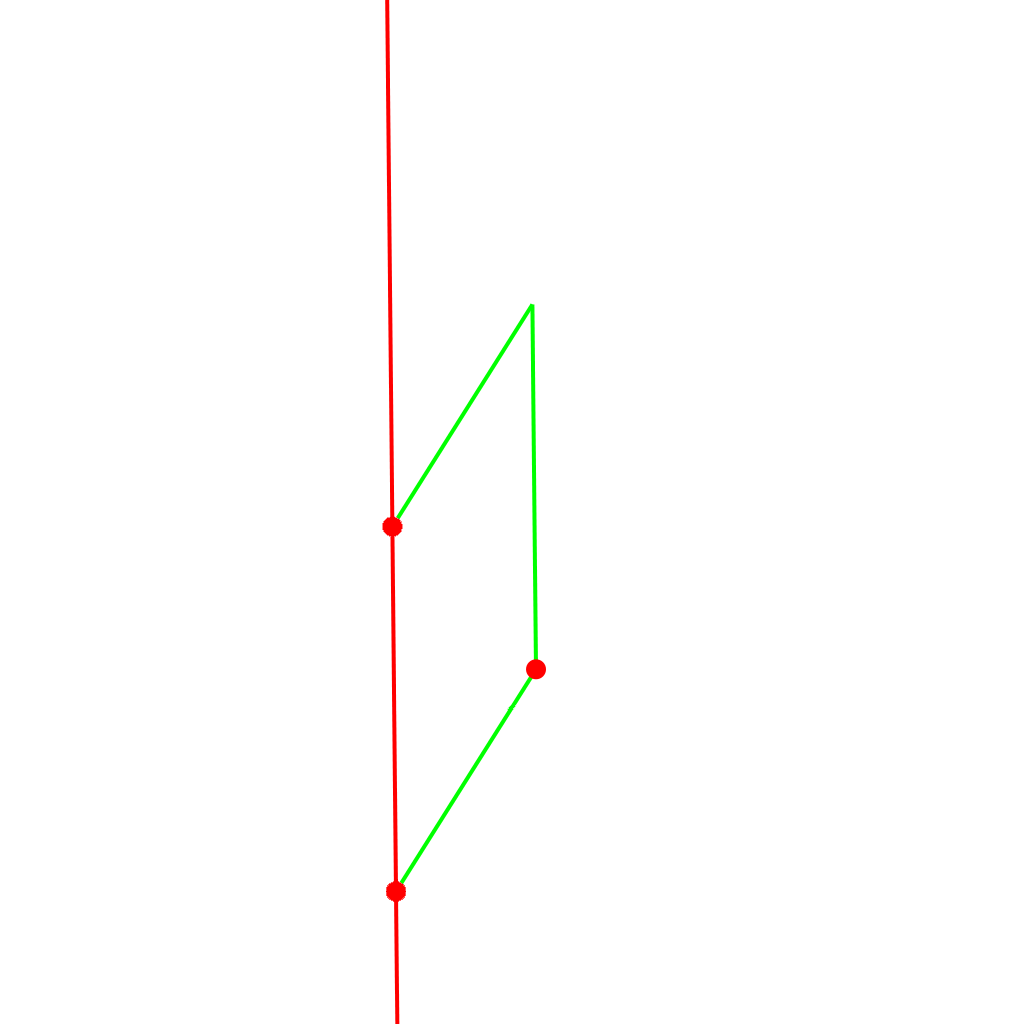}
         \caption{Current state}
         \label{fig:hypotheses_explorer_input}
     \end{subfigure}
     \hfill
     \begin{subfigure}[t]{0.32\textwidth}
         \centering
         \includegraphics[width=\textwidth]{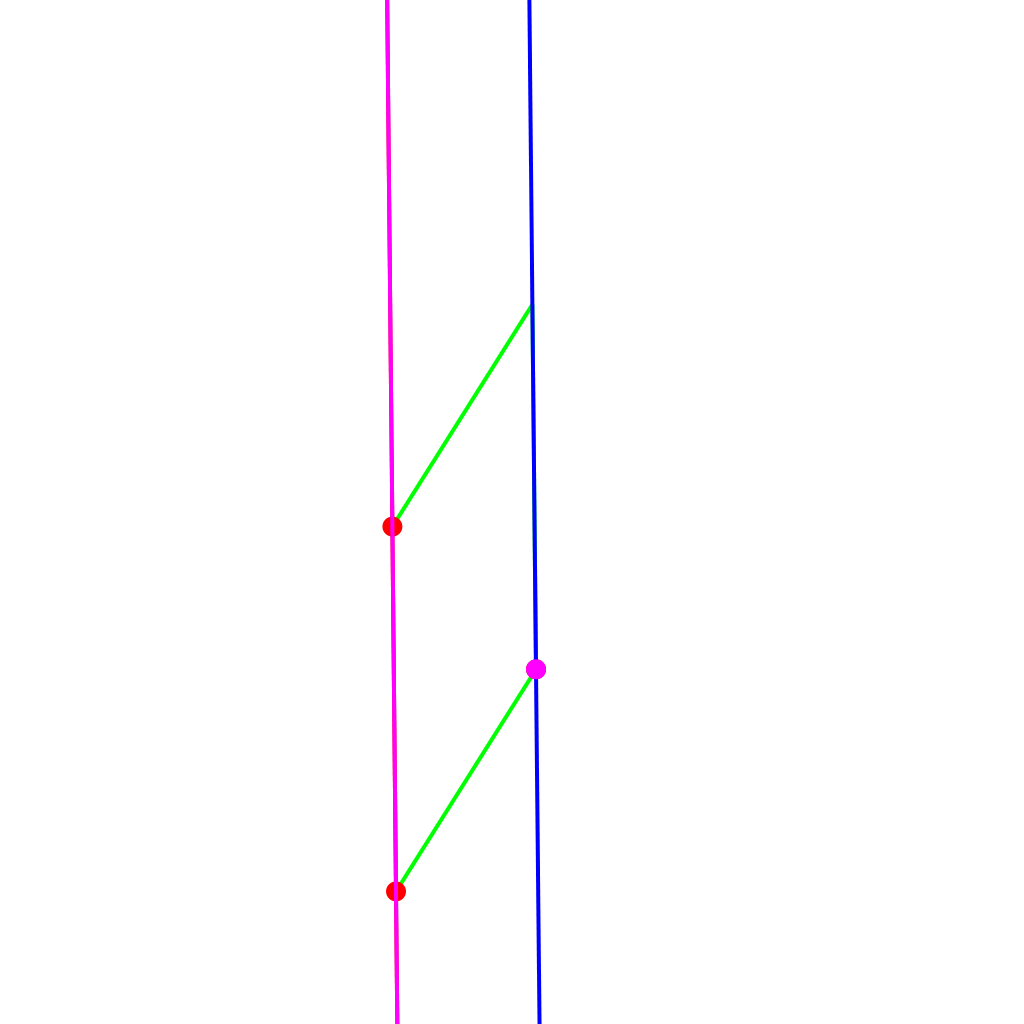}
         \caption{Hypothesis, reward $0.65$}
         \label{fig:hypotheses_explorer_hypothesis3}
     \end{subfigure}
     \hfill
     \begin{subfigure}[t]{0.32\textwidth}
         \centering
         \includegraphics[width=\textwidth]{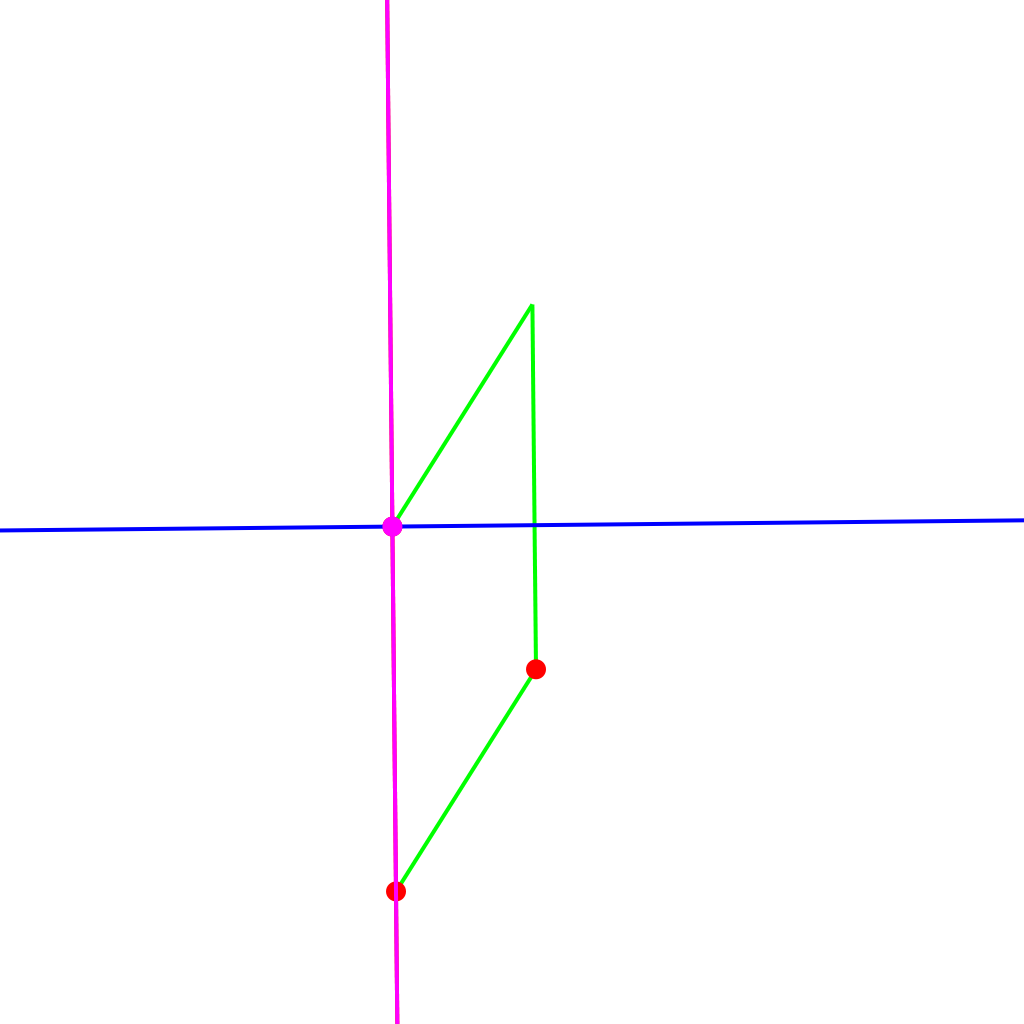}
         \caption{Hypothesis, reward $0$}
         \label{fig:hypotheses_explorer_hypothesis1}
     \end{subfigure}
     \hfill
     \begin{subfigure}[t]{0.32\textwidth}
         \centering
         \includegraphics[width=\textwidth]{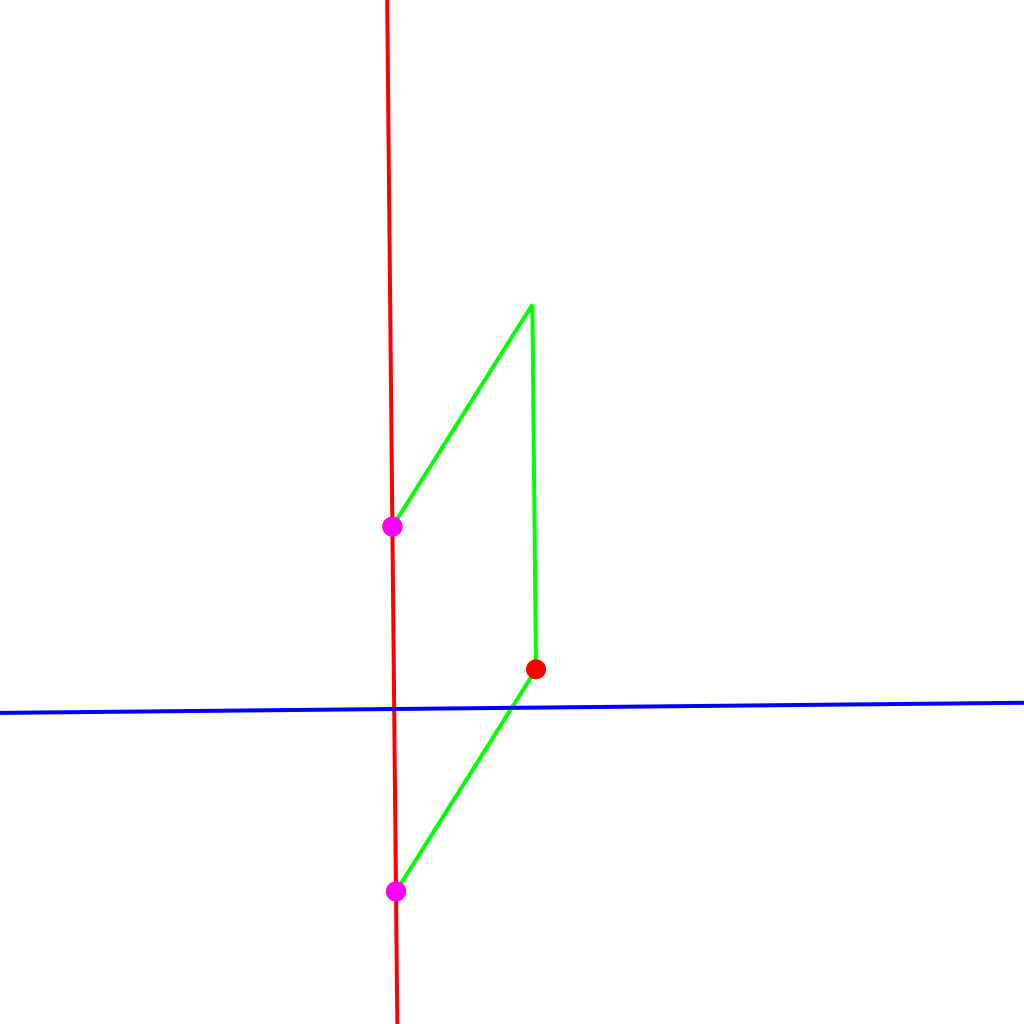}
         \caption{Hypothesis, reward $0$}
         \label{fig:hypotheses_explorer_hypothesis2}
     \end{subfigure}
     \hfill
     \begin{subfigure}[t]{0.32\textwidth}
         \centering
         \includegraphics[width=\textwidth]{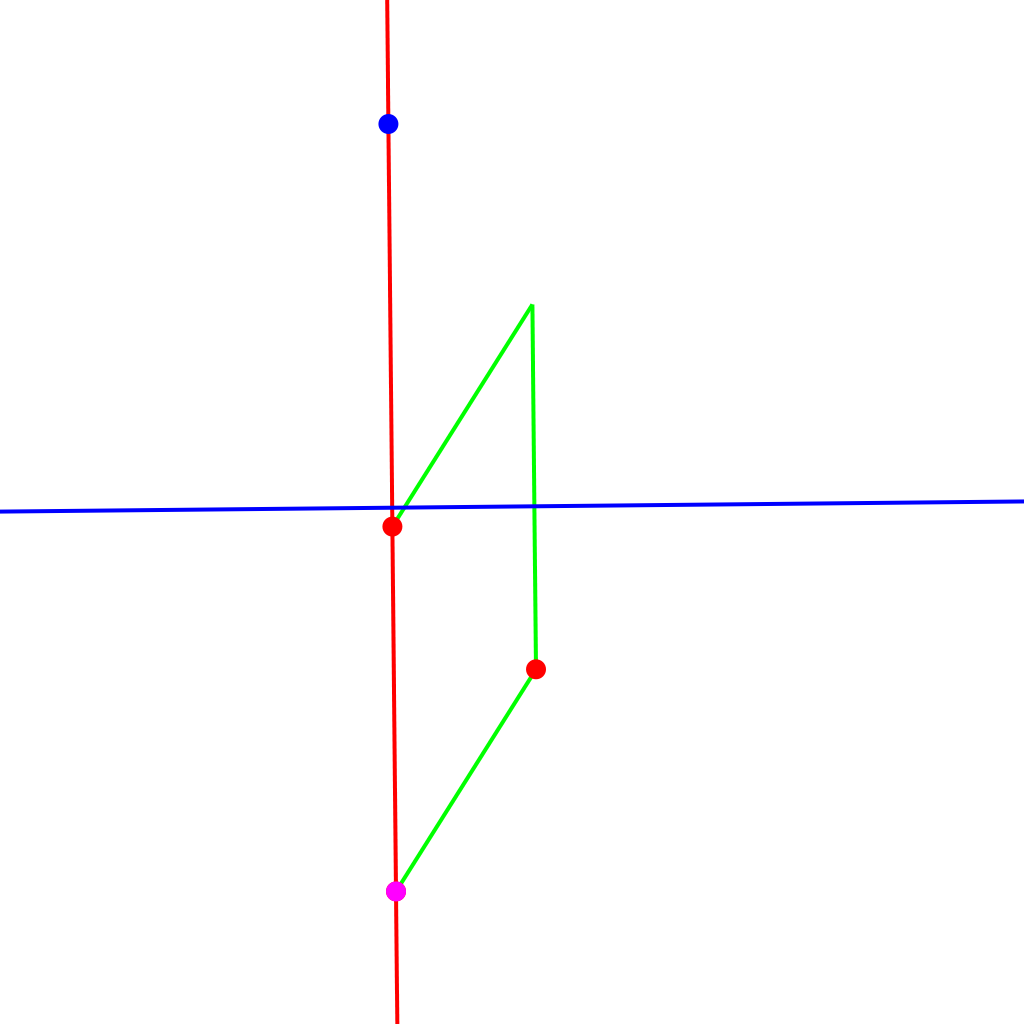}
         \caption{Hypothesis, reward $0$}
         \label{fig:hypotheses_explorer_hypothesis4}
     \end{subfigure}
     \hfill
     \begin{subfigure}[t]{0.32\textwidth}
         \centering
         \includegraphics[width=\textwidth]{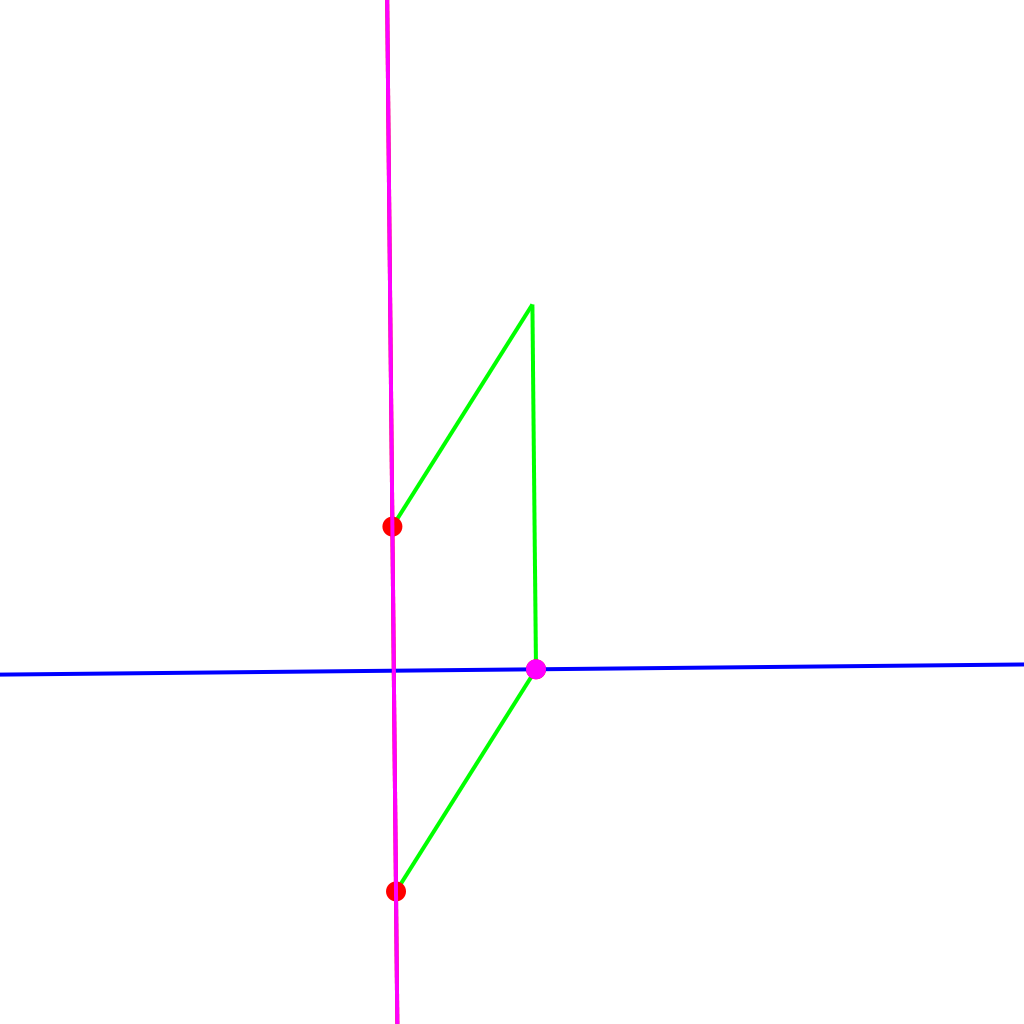}
         \caption{Hypothesis, reward $0$}
         \label{fig:hypotheses_explorer_hypothesis5}
     \end{subfigure}
        \caption{Multiple hypotheses for the next step in Euclidea level \textit{Epsilon-03} (construct a parallelogram whose three of four vertices are given). (a) Current state. (b) Hypothesis with reward $0.25$ that constructs one of the four lines in the goal (green). (c-f) Hypotheses with reward $0$. Each image contains the current state (red), remaining goal (green), hypothesis produced by \maskrcnn (blue), and
        parameters of the tool (purple). For example, hypothesis (b) selects the Parallel tool to construct the blue line as parallel to the purple line and intersecting the purple point. A positive reward indicates contribution to the goal, so we select the hypothesis (b), which has the highest reward.
        }
        \label{fig:hypotheses_explorer}
\end{figure}

Each primary detection from the \maskrcnn model (see Section \ref{position_dependent_pars}) can be transformed into an action.
We denote each action, its arguments, and results as a \emph{hypothesis}.
The result of an action contains a geometric primitive, constructed during the action execution, and a reward, indicating whether the output primitive is a part of the remaining goal or not.
If an action constructs a part of the goal, the reward is $1/n$, where $n$ is the total number of primitives in the initial goal, otherwise it is equal to zero.
Fig.~\ref{fig:hypotheses_explorer_hypothesis3} shows a hypothesis that successfully constructs one of the four lines in the goal and its reward thus equals $0.25$.

We can extract multiple actions from the \maskrcnn model output by considering multiple output candidate detections, transform them into multiple hypotheses, and explore their construction space.
We can also utilize hypotheses from models trained for different tasks. However, the \maskrcnn scores are not comparable across hypotheses from different models, so in a setup with multiple models we have to search even hypotheses with lower scores.
\subsection{Tree search for exploring construction hypotheses}
\label{hypotheses_tree_search}
We use tree search to explore the hypothesis space given by the predictions from one or more \maskrcnn models.
The tree search has to render the input image and obtain predictions from all considered \maskrcnn models in each node of the tree, which increases the time spent in one node.
However, as a result, we search only within the space of hypotheses output by the \maskrcnn models, which is much smaller than the space of all possible constructions.
We use iterative deepening, which is an iterated depth-limited search over increasing depth (see Algorithm~\ref{alg:IDDFS}).

\begin{small}
\begin{algorithm}[!htb]
\SetAlgoLined
\DontPrintSemicolon
\SetAlFnt{\small\sf} 
\KwResult{Solve level with Iterative Deepening.}

    \SetKwFunction{FMain}{IterativeDeepening-DFS}
    \SetKwProg{Fn}{Function}{:}{}
    \Fn{\FMain{$InitialState$}}{
        SolutionMaxLength $\longleftarrow 0$;
        
        Solution $\longleftarrow$ \textbf{null};
        
        \While{
        (Solution = \textbf{null}) and (SolutionMaxLength $<$ MaxIterationDepth)
        }
        {
        $SolutionMaxLength \longleftarrow SolutionMaxLength +1$;
        
        Solution $\longleftarrow$ FindSolution($InitialConfig$, SolutionMaxLength);
        
        }
        \textbf{return} Solution
}
\SetKwFunction{FMain}{FindSolution}
    \SetKwProg{Fn}{Function}{:}{}
    \Fn{\FMain{$CurrentState, Depth$}}{
        \If {CurrentState.IsTheGoal}
        {\textbf{return} success \tcp{collect solution on the backtracking}}
        \If{Depth $= 0$ }{\textbf{return null} \tcp{e.g.,~solution not found.}}
        
        $Hypotheses \longleftarrow Models.GetAllHypotheses(CurrentState)$;
        
        \tcp{Hypotheses sorted in the order: Reward, Confidence score}
        
        \ForEach{$h \in Hypotheses$}
            {
            $NewState \longleftarrow Apply(h, CurrentState)$;
            
            $Solution = FindSolution(NewState, Depth -1)$;
            
            \If {Solution $\neq$ \textbf{null}}
                {\textbf{return} Solution}
            }
        \textbf{return null}

}
\vspace{2mm}
\caption{\small{Tree search for exploring construction hypotheses using Iterative Deepening Depth-First Search algorithm.}}
\label{alg:IDDFS}
\end{algorithm}
\end{small}
\label{reducing_hypotheses}
Hypotheses produced by different \maskrcnn models increase the branching factor and thus also the search time.
To speed up the tree search, we group all mutually similar hypotheses and explore only one of them.
For this purpose, we consider two hypotheses as similar if their output geometric primitive is the same; note that such hypotheses can have different arguments, including different tools.

%% file: section_6_experiments.tex
\section{Experiments}
\label{experiments_section}
This section shows the performance of our method for both the {\em supervised setting}, where we see examples of the specific level in both training and test time, and the {\em unseen setting}, where we are testing on new levels, not seen during training.
We will compare the benefits of the different components of our approach and show an example solution produced by our method.

\subsection{Benefits of different components of our approach}
As the base method we use the off-the-shelf \maskrcnn approach (``Vanilla \maskrcnn''), which has a low accuracy even on simple \alphapack levels.
To improve over this baseline, we have introduced several additional components, namely, ``Automatic point detection'', ``History channel'', ``4+ Stage training'', ``Intersection degeneration rules'', and ``On-the-fly data generation'' (see Section~\ref{mrcnn_components}).
Table~\ref{components_performance} compares their cumulative benefits.
These components are crucial for solving levels in advanced level packs, e.g., the Gamma level pack, which could not be solved without the Intersection degeneration rules.

\begin{table}[!htb]
    \centering
    \begin{tabular}{p{48mm}|
    m{11mm}m{11mm}m{11mm}m{11mm}m{11mm}m{11mm}
    }
     Component / Level pack & \alphapack & \betapack & \gammapack & \deltapack & \epsilonpack & \zetapack  \\
     \hline
     Vanilla \maskrcnn~\cite{maskrcnn} & 71.0 & - & - & - & - & - \\
     + Automatic point detection & 95.1 & - & - & - & - &  - \\
     + History channel & 98.1 & 69.3 & - & - & - & -  \\ 
     + 4+ Stage training & 91.7 & 82.3 & - & - & - & - \\
     + Intersection degeneration rules & 91.7 & 82.3 & 79.9 & 73.5 &  - &  - \\
     + On-the-fly data generation & {\bf 98.7} & {\bf 96.2} & {\bf 97.8} & {\bf 99.1} & {\bf 92.8} & {\bf 95.7}\\
    \end{tabular}
    \caption{Performance of the base method (Vanilla \maskrcnn) together with the additional components of our approach. The Performance is evaluated on Euclidea level packs with increasing difficulty (from \alphapack to \zetapack). We trained a separate model for every level in each level pack and evaluated the model on 500 new  instances of that level.
    The table presents the accuracy averaged across all levels in each level pack.
    The components are applied to the base method (Vanilla \maskrcnn) in an additive manner.
    For example, the 4+ Stage training includes all previous components, i.e.,~Automatic point detection and History channel.
    A description of the different components is given in Section \ref{mrcnn_components}.}
    \label{components_performance}
    \vspace{-3.3em}
\end{table}

\begin{figure}[!htb]
    \centering
    \includegraphics[width=0.88\textwidth]{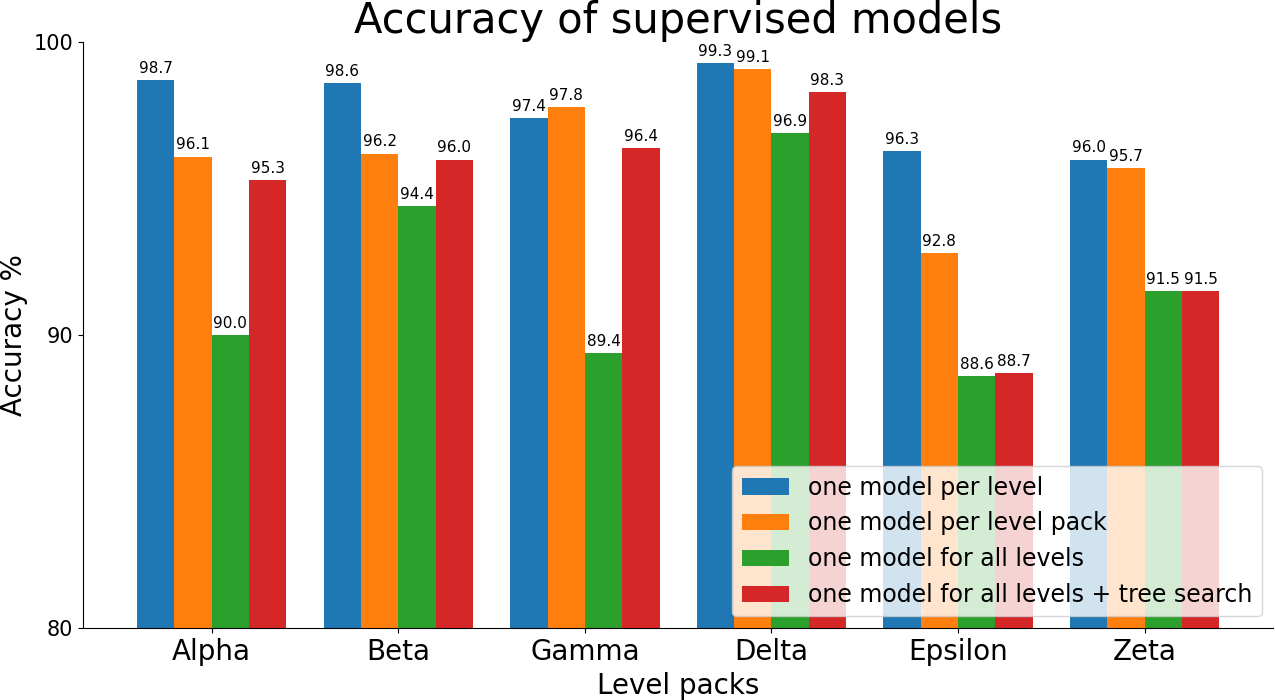}
    \caption{Accuracy of our approach on Euclidea level packs Alpha to Zeta.
    The model uses all components from Table \ref{components_performance}.
    We compare four approaches: one model per level (blue), one model per level pack (orange), one model for all levels (green), and one model for all levels with hypothesis tree search (red). All were evaluated on 500 instances of each level and the accuracy was averaged across all levels in each level pack.
    }
    \label{final_results}
    \vspace{-2em}
\end{figure}
\subsection{Evaluation of the supervised learning approach}

We evaluated our method on the first six level packs of Euclidea with various training setups. The results (see Fig.~\ref{final_results}) show that models specialized to individual level packs (``one model per level pack'') or even levels (``one model per level'') have better accuracy than a single, more general model for multiple levels/packs (``one model for all levels''). We also investigate the benefits of using our tree search procedure (see Section~\ref{hypotheses_tree_search}) instead of using only the most confident hypothesis, as in the supervised setting. We find the tree search improves the accuracy, especially on \alphapack and \gammapack level packs, by searching the space of possible candidate solutions for each step of the construction.
\subsection{Evaluation on unseen problems}

To evaluate performance on unseen levels, we use the leave-one-out (LOO) evaluation.
In our setup, \emph{LOO levels} evaluates, e.g., level \textit{Alpha-01} using models trained on each of the other \alphapack levels, whereas \emph{LOO packs} evaluates, e.g., each \alphapack level using models trained on level packs \betapack, \gammapack, \deltapack, \epsilonpack, and \zetapack.
Table~\ref{tab:loo_eval} compares the LOO evaluation and the supervised approach, both with the hypothesis tree search, for level pack \alphapack.
We ran a similar evaluation for all 6 levels packs and were able to solve 30 out of 68 levels using \emph{LOO levels} and 31 out of 68 levels using \emph{LOO packs}.
The results show that our method can solve many levels unseen during the training, although some levels remain difficult to solve.
\label{eval_of_unseen_levels}
\begin{table}[!htb]
    \centering
    \setlength{\tabcolsep}{12pt}
    \begin{tabular}{l|ccc}
     \alphapack levels & LOO levels & LOO packs& supervised\\
     \hline
     01 T1 Line & 40.0 & 10.0 & 85.0\\
     02 T2 Circle & 5.0 & 45.0 & 100.0\\
     03 T3 Point & 100.0 & 90.0 & 100.0\\
     04 TIntersect & 100.0 & 100.0 & 99.0\\
     05 TEquilateral & 50.0 & 70.0 & 100.0\\
     06 Angle60 & 55.0 & 100.0 & 94.0\\
     07 PerpBisector & 35.0 & 100.0 & 99.0\\
     08 TPerpBisector & 0.0 & 100.0 & 75.0\\
     09 MidPoint & 5.0 & 60.0 & 100.0\\
     10 CircleInSquare & 0.0 & 100.0 & 87.0\\
     11 RhombusInRect & 25.0 & 40.0 & 99.0\\
     12 CircleCenter & 10.0 & 0.0 & 100.0\\
     13 SquareInCircle & 0.0 & 10.0 & 100.0\\
     \hline
     Average & 42.5 & 63.4 &95.3\\
\end{tabular}
    \caption{Leave-one-out evaluation on the Alpha levels.
    Completion accuracy of the leave-one-out evaluation across levels within a level pack (LOO levels), across level packs (LOO packs), and, for comparison, our best supervised model trained to solve the first six Euclidea level packs (\alphapack-\zetapack); the tree search was used in all three cases.
    The leave-one-out evaluation was performed on 20 instances of each level, while the supervised model was evaluated on 500 instances of each level. Using models from all level packs except Alpha (LOO packs) works better than using models trained only on other levels of the Alpha level pack (LOO levels). This is likely because models in the ``LOO packs" set-up have seen a larger set of different constructions.
}
    \label{tab:loo_eval}
    \vspace{-2em}
\end{table}

\newpage
\subsection{Qualitative example}

\begin{figure}[!h]
     \centering
     \begin{subfigure}[t]{0.32\textwidth}
         \centering
         \includegraphics[width=\textwidth]{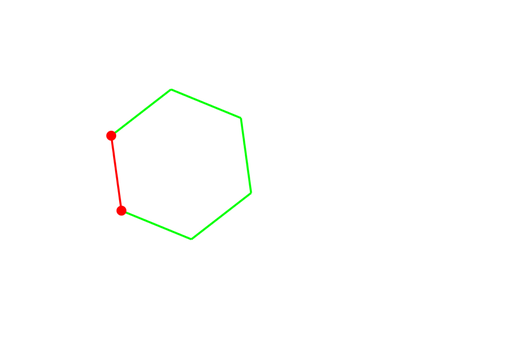}
         \caption{Input}
         \label{fig:Epsilo12_example_input}
     \end{subfigure}
     \hfill
     \begin{subfigure}[t]{0.32\textwidth}
         \centering
         \includegraphics[width=\textwidth]{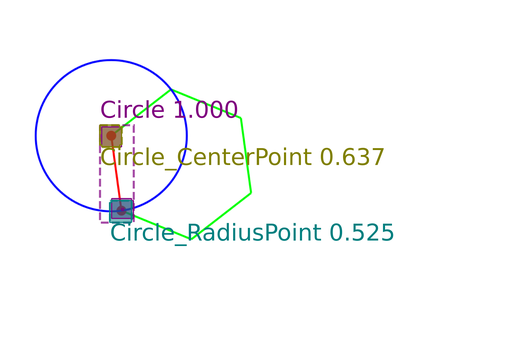}
         \caption{Step 1: Circle tool}
         \label{fig:Epsilon12_example_step1}
     \end{subfigure}
     \hfill
     \begin{subfigure}[t]{0.32\textwidth}
         \centering
         \includegraphics[width=\textwidth]{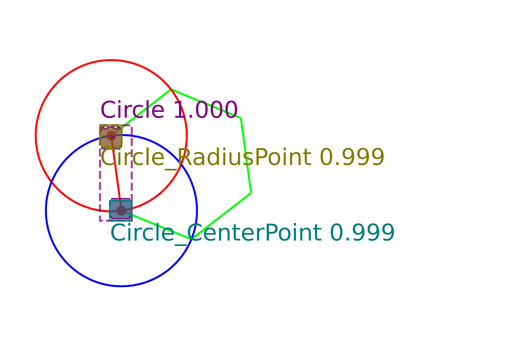}
         \caption{Step 2: Circle tool}
         \label{fig:Epsilon12_example_step2}
     \end{subfigure}
     \hfill
     \begin{subfigure}[t]{0.32\textwidth}
         \centering
         \includegraphics[width=\textwidth]{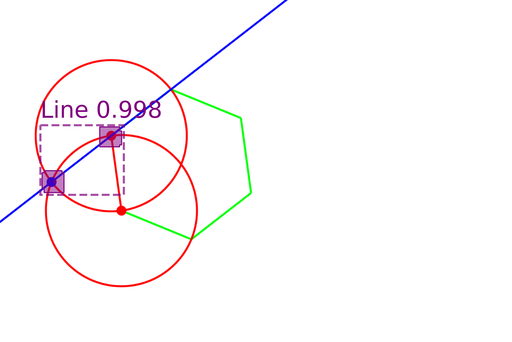}
         \caption{Step 3: Line tool}
         \label{fig:Epsilon12_example_step3}
     \end{subfigure}
     \hfill
     \begin{subfigure}[t]{0.32\textwidth}
         \centering
         \includegraphics[width=\textwidth]{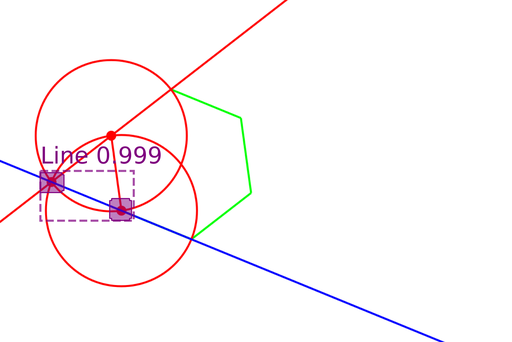}
         \caption{Step 4: Line tool}
         \label{fig:Epsilon12_example_step4}
     \end{subfigure}
     \hfill
     \begin{subfigure}[t]{0.32\textwidth}
         \centering
         \includegraphics[width=\textwidth]{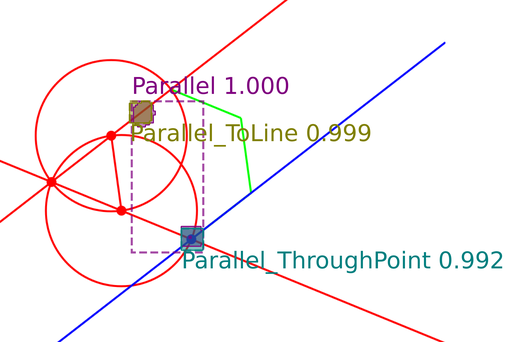}
         \caption{Step 5: Parallel tool}
         \label{fig:Epsilon12_example_step5}
     \end{subfigure}
     \hfill
     \begin{subfigure}[t]{0.32\textwidth}
         \centering
         \includegraphics[width=\textwidth]{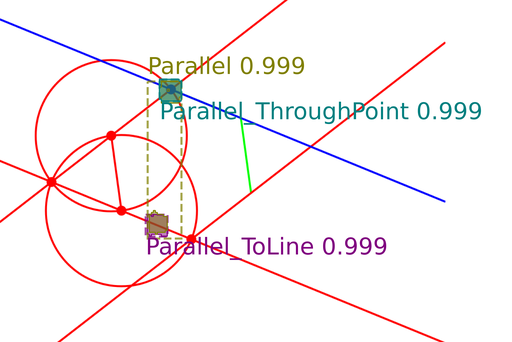}
         \caption{Step 6: Parallel tool}
         \label{fig:Epsilon12_example_step6}
     \end{subfigure}
     \hfill
     \begin{subfigure}[t]{0.32\textwidth}
         \centering
         \includegraphics[width=\textwidth]{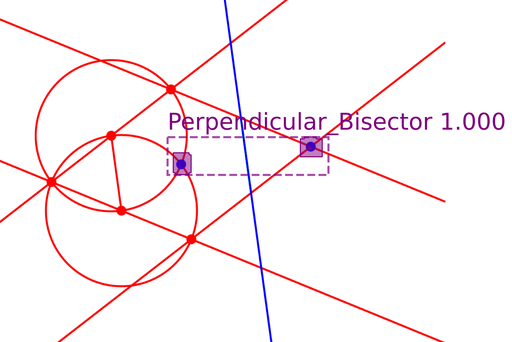}
         \caption{Step 7: Perpendicular tool}
         \label{fig:Epsilon12_example_step7}
     \end{subfigure}
     \hfill
     \begin{subfigure}[t]{0.32\textwidth}
         \centering
         \includegraphics[width=\textwidth]{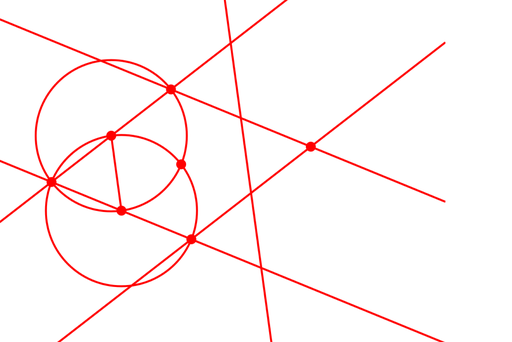}
         \caption{Construction finished}
         \label{fig:Epsilon12_example_fin}
     \end{subfigure}
     \hfill
     
        \caption{Example construction of Euclidea level \textit{Epsilon-12} (construct a regular hexagon by the side).
        (a) Initial configuration of the problem.
        (b-h) Seven construction steps, including \maskrcnn detections, and an object proposed to construct in each step.
        (i) Final construction step, level solved.
        Red denotes the current state of the construction, green the remaining goal,
        blue the geometric primitive proposed by the detection,
        and other colors
        the prediction masks, bounding boxes, class names, and scores for the next predicted action.
        More examples can be found in the supplementary material available at the project webpage~\cite{project-page}.
        }
        \label{fig:Epsilon12_example}
\end{figure}

Fig.~\ref{fig:Epsilon12_example} shows a step-by-step walk-through construction of an advanced Euclidea level.

\begin{figure}[htb!]
    \centering
    \includegraphics[width=0.4\textwidth]{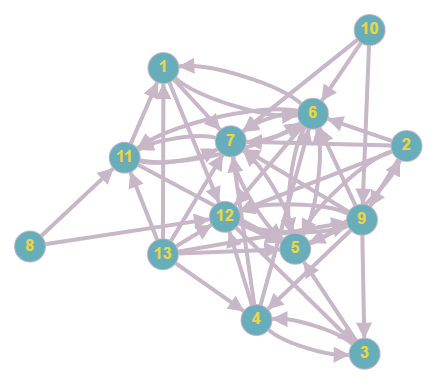}
    \caption{
    \small Connection graph created during the leave-one-out evaluation of the Alpha level pack using models for individual levels. Models trained on 
    more difficult levels (indicated by higher numbers) often help construct easier levels (lower numbers). For example, level 13 could not be solved (has no incoming connections), but its predictions were used for solution of levels 1, 4, 5, 7, 9, 11, 12. Our method can often construct simpler tasks based on more complex constructions. However, combining multiple simple tasks into a complex construction remains a challenge.
    }
    \label{connection_graph}
\end{figure}

\paragraph{\textbf{Connections between levels.}}
\label{connections}
From the leave-one-out evaluation, we can also observe which levels are similar.
We denote that level $X$ is \emph{connected} to level $Y$ if a model trained for $Y$ contributes with a hypothesis to a successful construction during the inference for level $X$.
Note that this relation is not symmetric, e.g.,~when $X$ is connected to $Y$, then $Y$ is not necessarily connected to $X$.
We run the hypothesis tree search during the leave-one-out evaluation and obtain connections in the following way: If the search is successful, we collect all models that contributed to the solution in the final backtracking of the search.
The connections for all levels in the level pack Alpha are shown in Fig.~\ref{connection_graph}.

%% file: section_7_conclusion.tex
\section{Conclusion}

We have developed an image-based method for solving geometric construction problems. The method builds on the \maskrcnn visual recognizer, which is adapted to predict the next step of a geometric construction given the current state of the construction, and further combined with a tree search mechanism to explore the space of possible constructions. To train and test the method, we have used Euclidea, a construction game with geometric problems with an increasing difficulty. To train the model, we have created a data generator that generates new configurations of the Euclidea constructions. 

In a supervised setting, the method learns to solve all 68 kinds of geometric construction problems from the first six level packs of Euclidea with an average 92\% accuracy. When evaluated on new kinds of problems unseen at training, which is a significantly more challenging set-up, our method solves 31 of the 68 kinds of Euclidea problems. 
To solve the unseen problems our model currently relies on having seen a similar (or more complex) problem at training time. Solving unseen problems that are more complex than those seen at training remains an open challenge. Addressing this challenge is likely going to require developing new techniques to efficiently explore the space of constructions as well as mechanisms to learn from successfully completed constructions, a set-up similar to reinforcement learning. 

Although in this paper we focus on synthetically generated data, our results open up the possibility to solve even hand-drawn geometric problems if corresponding training data is provided. 
In real-world settings, descriptions and discussions of problems in math, physics and other sciences often contain an image or a drawing component. Diagrams and illustrations are also an essential part of real-world technical documentation (e.g.~patents). The ability to automatically identify and combine visually defined geometrical primitives into more complex patterns, as done in our work, is a step towards automatic systems that can identify compositions of geometric patterns in technical documentation. Think, for example, of an ``automatic patent lawyer assistant''.